\documentclass[conference]{IEEEtran}

\IEEEoverridecommandlockouts                              

\usepackage{microtype}
\usepackage{amsmath,amssymb,amsfonts}
\usepackage{algorithmic}
\usepackage{graphicx}
\usepackage{textcomp}
\usepackage{xcolor}
\usepackage{subfig}
\usepackage{gensymb}
\usepackage{caption}
\usepackage{hyperref}
\usepackage{mwe}
\usepackage{amsfonts}
\usepackage{amssymb}
\usepackage{verbatim}
\usepackage{amsbsy}
\usepackage{siunitx}
\usepackage{tabularx, booktabs}
\usepackage{dblfloatfix}
\usepackage{cite}
\usepackage{cleveref}

\usepackage{algorithm, algorithmic}
\usepackage[autolanguage]{numprint}

\usepackage{spreadtab}
\usepackage{multirow}
\usepackage{cite}

\usepackage{hyperref}
\usepackage{xcolor}
\usepackage{algorithmic}
\usepackage{textcomp}
\usepackage{xcolor}
\usepackage{subfig}
\usepackage{gensymb}
\usepackage{caption}

\usepackage{mwe}
\usepackage{amsfonts}
\usepackage{amssymb}
\usepackage{verbatim}
\usepackage{amsbsy}
\usepackage{siunitx}
\usepackage{tabularx, booktabs}
\usepackage{soul}
\usepackage{tikz}
\usetikzlibrary{calc}

\makeatletter
\newif\if@anonymize

\@anonymizetrue    

\if@anonymize
  \newcommand{\highlight@DoHighlight}{
    \fill [outer sep = -15pt, inner sep = 0pt, color=black]
          ($(begin highlight)+(0,8pt)$) rectangle ($(end highlight)+(0,-3pt)$) ;
  }

  \newcommand{\highlight@BeginHighlight}{
    \coordinate (begin highlight) at (0,0) ;
  }

  \newcommand{\highlight@EndHighlight}{
    \coordinate (end highlight) at (0,0) ;
  }

  \newdimen\highlight@previous
  \newdimen\highlight@current
  \newlength{\item@width}

  \DeclareRobustCommand*\anonymize{%
    \SOUL@setup
    \def\SOUL@preamble{%
      \begin{tikzpicture}[overlay, remember picture]
        \highlight@BeginHighlight
        \highlight@EndHighlight
      \end{tikzpicture}%
    }%
    \def\SOUL@postamble{%
      \begin{tikzpicture}[overlay, remember picture]
        \highlight@EndHighlight
        \highlight@DoHighlight
      \end{tikzpicture}%
    }%
    \def\SOUL@everyhyphen{%
      \discretionary{%
        \SOUL@setkern\SOUL@hyphkern
        \SOUL@sethyphenchar
        \tikz[overlay, remember picture] \highlight@EndHighlight ;%
      }{%
      }{%
        \SOUL@setkern\SOUL@charkern
      }%
    }%
    \def\SOUL@everyexhyphen##1{%
      \SOUL@setkern\SOUL@hyphkern
      \settowidth{\item@width}{##1}%
      \makebox[\item@width]{}%
      \discretionary{%
        \tikz[overlay, remember picture] \highlight@EndHighlight ;%
      }{%
      }{%
        \SOUL@setkern\SOUL@charkern
      }%
    }%
    \def\SOUL@everysyllable{%
      \begin{tikzpicture}[overlay, remember picture]
        \path let \p0 = (begin highlight), \p1 = (0,0) in \pgfextra
          \global\highlight@previous=\y0
          \global\highlight@current =\y1
        \endpgfextra (0,0) ;
        \ifdim\highlight@current < \highlight@previous
          \highlight@DoHighlight
          \highlight@BeginHighlight
        \fi
      \end{tikzpicture}%
      \settowidth{\item@width}{\the\SOUL@syllable}%
      \makebox[\item@width]{}%
      \tikz[overlay, remember picture] \highlight@EndHighlight ;%
    }%
    \SOUL@
  }
\else
  \newcommand{\anonymize}[1]{#1}
\fi

\makeatother 

\title{\LARGE \bf
DoCRL: Double Critic Deep Reinforcement Learning for Mapless Navigation of a Hybrid Aerial Underwater Vehicle with Medium Transition}

\author{\IEEEauthorblockN{1\textsuperscript{st} Ricardo B. Grando}
\IEEEauthorblockA{\textit{ Robotics Lab} \\
\textit{ Technological University of Uruguay}\\
 Rivera, Uruguay \\
 ricardo.bedin@utec.edu.uy}
\and
\IEEEauthorblockN{2\textsuperscript{nd}  Junior C. de Jesus}
\IEEEauthorblockA{\textit{ Centro de Ci\^encias Computacionais} \\
\textit{ Universidade Federal de Rio Grande}\\
 Rio Grande, Brazil \\
 dranaju@gmail.com}
\and
\IEEEauthorblockN{3\textsuperscript{rd}   Victor A. Kich}
\IEEEauthorblockA{\textit{ Universidade Federal de Santa Maria} \\
\textit{ Universidade Federal de Santa Maria (UFSM)}\\
 Santa Maria, Brazil \\
 victorkich@yahoo.com.br}
\and
\IEEEauthorblockN{4\textsuperscript{th}  Alisson H. Kolling}
\IEEEauthorblockA{\textit{ Universidade Federal de Santa Maria} \\
\textit{ Universidade Federal de Santa Maria (UFSM)}\\
 Santa Maria, Brazil \\
 alikolling@gmail.com }
\and
\IEEEauthorblockN{5\textsuperscript{th}  Rodrigo S. Guerra}
\IEEEauthorblockA{\textit{ Centro de Ci\^encias Computacionais} \\
\textit{ Universidade Federal de Rio Grande}\\
 Rio Grande, Brazil \\
 rodrigo.guerra@furg.br }
\and
\IEEEauthorblockN{6\textsuperscript{th}  Paulo L. J. Drews-Jr}
\IEEEauthorblockA{\textit{ Centro de Ci\^encias Computacionais} \\
\textit{ Universidade Federal de Rio Grande}\\
 Rio Grande, Brazil \\
 paulodrews@furg.br}
}

\begin{document}

\maketitle
\thispagestyle{empty}
\pagestyle{empty}

\begin{abstract}
Deep Reinforcement Learning (Deep-RL) techniques for motion control have been continuously used to deal with decision-making problems for a wide variety of robots. Previous works showed that Deep-RL can be applied to perform mapless navigation, including the medium transition of Hybrid Unmanned Aerial Underwater Vehicles (HUAUVs). These are robots that can operate in both air and water media, with future potential for rescue tasks in robotics. This paper presents new approaches based on the state-of-the-art Double Critic Actor-Critic algorithms to address the navigation and medium transition problems for a HUAUV. We show that double-critic Deep-RL with Recurrent Neural Networks using range data and relative localization solely improves the navigation performance of HUAUVs. Our DoCRL approaches achieved better navigation and transitioning capability, outperforming previous approaches. 
\end{abstract}


\vspace{-3mm}
\section*{Supplementary Material}\label{supplementary_material}

Video of the experiments available at: \texttt{\url{https://youtu.be/PqTDzsKjA9c}}. 
Released code at: \texttt{\url{https://github.com/ricardoGrando/DoCRL}}.

\vspace{-2.5mm}
\section{Introduction}
\label{introduction}


Several studies about Hybrid Unmanned Aerial Underwater Vehicles (HUAUVs) have been conducted recently \cite{drews2014hybrid, neto2015attitude, da2018comparative, maia2017design, lu2019multimodal, mercado2019aerial,horn20,aoki2021}. These vehicles provide an interesting range of possible applications due to the capability to act in two different environments, including inspection and mapping of partly submerged areas in industrial facilities, search and rescue and other military-related applications. However, the state-of-the-art is yet focused on the vehicle design and structure, where even fewer studies around autonomous navigation have been conducted \cite{bedin2021deep}. The ability to perform tasks in both environments and successfully transit between them imposes additional challenges that must be addressed to make this mobile vehicle autonomously feasible. 

\begin{figure}[tbp!]
    \vspace{-2mm}
    \centering
    \includegraphics[width=\linewidth]{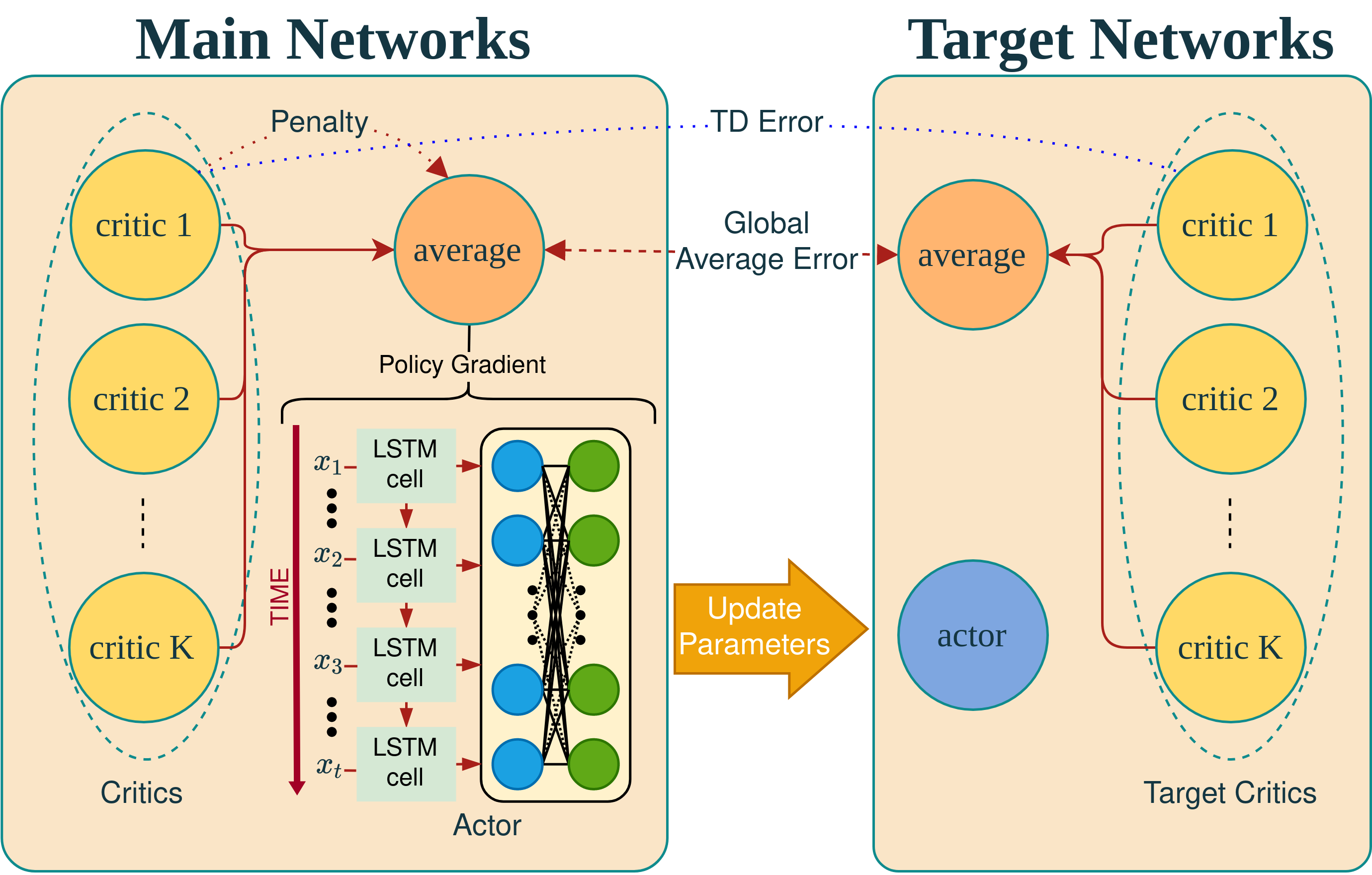}
    \caption{DoCRL architecture.}
    \label{fig:system_def}
    \vspace{-4mm}
\end{figure}

Approaches based on Deep Reinforcement Learning (Deep-RL) techniques have been enhanced to address navigation-related tasks for a range of mobile vehicles, including ground mobile robots \cite{ota2020efficient}, aerial robots \cite{tong2021uav,grando2022double} and underwater robots \cite{carlucho2018}. These approaches based on single critic actor-critic techniques with multi-layer network structures have achieved interesting results in performing mapless navigation, obstacle avoidance and media transitioning even for HUAUVs \cite{bedin2021deep}. However, the challenges faced by this kind of vehicle make these approaches limited, not being capable of escalating to more complex scenarios for a rescue navigation task, for example.


In this work, we explore the use of Deep-RL in the context of HUAUVs to perform navigation-related tasks that can simulate through environmental rescue tasks in robotics. We present two enhanced approaches based on state-of-the-art Deep-RL for the continuous state:  (1) a deterministic based on Twin Delayed Deep Deterministic Policy Gradient (TD3) \cite{fujimoto2018addressing}; and
 (2) a stochastic based on Soft Actor-Critic (SAC) \cite{haarnoja2018soft}. We show that we are capable of training agents with a consistently better capability than state-of-the-art, performing with more stability the mapless navigation, obstacle avoidance and medium transition. We perform a two-fold way evaluation with air-to-water and water-to-air navigation. We compare our DoCRL approaches with single critic-based approaches used to perform mapless navigation and with an adapted version of a traditional Behavior-Based Algorithm (BBA) \cite{marino2016minimalistic} used in aerial vehicles. Our proposed double critic formulation can be seen in Fig. \ref{fig:system_def}.
 




This work contains the following main contributions:

\begin{itemize}

\item We propose two approaches based on state-of-the-art actor-critic double critic Deep-RL algorithms that can successfully perform goal-oriented mapless navigation for HUAUVs, using only range data readings and the vehicles' relative localization data.

\item We show that a Long Short Term Memory (LSTM) architecture achieves better overall performance than the state-of-the-art Multi-Layer Perceptron (MLP) architecture.


\item We show that our robot presents a robust capacity to navigate in scenarios that can simulate through environmental (air-water) rescue tasks in robotics. The robot also performs the medium transition, capable of arriving at the desired target and avoiding collisions.


\end{itemize}

This work has the following structure: the related works are discussed in the following section (Sec. \ref{related_works}). Following it, we present our methodology in Sec. \ref{methodology}. The results are presented in Sec. \ref{results} and discussed in Sec. \ref{discussion}. Finally, we discuss our contributions and present future works in Sec. \ref{conclusion}.

\section{Related Work}
\label{related_works}

The HUAUV literature is still mostly concerned with mechanical design and modelling \cite{drews2014hybrid, neto2015attitude, da2018comparative, maia2017design, lu2019multimodal, mercado2019aerial, horn20}. Autonomous navigation-related problems are still barely addressed for this kind of vehicle \cite{bedin2021deep}. The HUAUV used in this paper \cite{bedin2021deep} was created based on Drews-Jr \emph{et al.} \cite{drews2014hybrid} model, which Neto \emph{et al.} \cite{neto2015attitude} has largely expanded.


Several Deep-RL works in robotics have previously been carried out for the mapless navigation problem, demonstrating how efficiently we may solve the problem utilizing learning techniques
~\cite{tobin2017domain}. For a mapless motion planner of a ground robot, Tai \emph{et al.} \cite{tai2017virtual} employed 10-dimensional range findings and the relative distance of the vehicle to a target as inputs and continuous steering signals as outputs. According to the results, a mapless motion planner based on the DDPG algorithm may be effectively taught and utilized to navigate to a target. Recently, deep-RL methods have also been successfully used in robotics by 
Ota \emph{et al.} \cite{ota2020efficient}, de Jesus \emph{et al.}. \cite{jesus2019deep,jesus2021soft}, and others to accomplish mapless navigation-related tasks for terrestrial mobile robots. 


Kelchtermans and Tuytelaars \cite{kelchtermans2017hard} used a LSTM to perform autonomous navigation in a UAV. The room crossing task performed by Kelchtermans and Tuytelaars approach demonstrated through simulation how memory can help Deep Neural Networks (DNN) in navigation control. Tong \emph{et al.} \cite{tong2021uav} proposed a DRL-based method using a LSTM to navigate a UAV in high dynamic environments, with numerous obstacles moving fast. Their approach achieved superiority in terms of convergence and effectiveness compared with the state-of-the-art DRL methods. 
Singh and Thongam \cite{singh2018mobile} employed a Multi-Layer Perceptron (MLP) to perform terrestrial mobile robot navigation in dynamic environments. Their method is used to choose a collision-free segment and controls the robot's speed for each motion. They demonstrated that the method is efficient and gives a near-optimal path reaching the target position of the mobile robot.


When it comes to problems involving mapless navigation for Unmanned Aerial Vehicles, the effectiveness of Deep-RL is somewhat limited (UAVs) \cite{grando2020visual}. Rodriguez \emph{et al.} \cite{rodriguez2018deep} employed a DDPG-based strategy to solve the problem of landing on a moving platform. It employed Deep-RL in conjunction with the RotorS framework \cite{furrer2016rotors} to simulate UAVs in the Gazebo simulator. Sampedro \emph{et al.} \cite{sampedro2019fully} proposed a DDPG-based strategy for the Search and Rescue mission in interior situations, utilizing visual data from a real and simulated UAV. Kang \emph{et al.} \cite{kang2019generalization} also used visual information, although he focused on the subject of collision avoidance. In a go-to-target task, Barros \emph{et al.} \cite{2020arXiv201002293M} applied a SAC-based method to low-level control of a UAV. Grando \emph{et al.} \cite{grando2020deep} utilized approaches based on the DDPG and SAC algorithms on Gazebo for 2D UAV navigation. Recently, double critic-based Deep-RL approaches have been developed for UAVs, presenting better results \cite{grando2022double}.



Two works have recently tackled the navigation problem with the medium transition of HUAUVs \cite{pinheiro2021trajectory}, \cite{bedin2021deep}. Pinheiro \emph{et al.} \cite{pinheiro2021trajectory} focused on smoothing the medium transition problem in a simulated model on MATLAB. Grando \emph{et al.} \cite{bedin2021deep} developed Deep-RL approaches following single critic structure and a MLP architecture. These two works were developed using generic distance sensing information for aerial and underwater navigation. In contrast, more sophisticated sensing with a real-world simulated LIDAR and sonar is adopted in the present work.

Our work differs from the previously discussed works by only using the vehicle's relative localization data and not its explicit localization data. We also present Deep-RL approaches based on double critic techniques instead of single critic, with RNN structures instead of MLP, traditionally used for mapless navigation of mobile robots. We compare our DoCRL approaches with state-of-the-art Deep-RL approaches and with a behavior-based algorithm \cite{marino2016minimalistic} adapted for hybrid vehicles to show that our methodology improves the overall capability to perform autonomous navigation.

\section{Methodology}
\label{methodology}

In this section, we discuss our Deep-RL approaches. We detail the network structure for both deterministic and stochastic agents. We also present the proposed reward function for the task that the vehicle must accomplish autonomously.




\subsection{Double Critic Deep Reinforcement Learning Deterministic - DoCRL-D}

Developing on the DQN \cite{mnih2013playing}, Deep Deterministic Policy Gradient (DDPG)~\cite{lillicrap2015continuous} employs an actor-network where output is a real value that represents a chosen action, and a second neural network to learn the target function providing stability and making it ideal for mobile robots~\cite{jesus2019deep}. While providing good results, DDPG still has its problems, like overestimating the Q-values leading to policy breaking. TD3 \cite{fujimoto2018addressing} uses DDPG as its backbone and improves it by adding some improvements, such as clipped double-Q learning with two neural networks as targets for the Bellman error loss functions, delayed policy updates, and Gaussian noise on the target action, raising its performance.    
Our deterministic approach is based on the TD3 technique. The pseudocode of DoCRL-D can be seen in Algorithm \ref{alg:docrl_d}.

\begin{algorithm}[!htb]
    \algsetup{linenosize=\tiny}
    \scriptsize
    \caption{Double Critic Deep Reinforcement Learning Deterministic - DoCRL-D}
    \label{alg:docrl_d}
    \begin{algorithmic}[1]
        \STATE Initialize params of critic networks $\theta_{1}$, $\theta_{2}$ , and actor network $\pi(\phi)$
        \STATE Initialize params of target networks $\phi^{\prime}\leftarrow\phi$, $\theta_{1}^{\prime}\leftarrow\theta_{1}$, $\theta_{2}^{\prime}\leftarrow\theta_{2}$
        \STATE Initialize replay buffer $\beta$
        \FOR{$ep = 1$ to $max\_eps$}
            \STATE reset environment state
            \FOR{$t = 0$ to $max\_steps$}
                \IF {$t < start\_steps$}
                    \STATE $a_{t} \leftarrow $ env.action\_space.sample() 
                \ELSE
                    \STATE $a_{t}\leftarrow\pi_{\phi}(s)+\epsilon,\ \epsilon\sim \mathcal{N}(0,OU)$
                \ENDIF
                
                \STATE $s_{t+1}$, $r_{t}$, $d_{t}$, \_ $\leftarrow$ env.step($a_{t}$)
                
                \STATE store the new transition $(s_{t}, a_{t}, r_{t}, s_{t+1}, d_{t})$ into $\beta$
                
                \IF{$t > start\_steps$}
                    \STATE Sample mini-batch of $N$ transitions $(s_{t}, a_{t}, r_{t},s_{t+1}, d_{t})$ from $\beta$
                    
                    \STATE $a'\leftarrow\pi_{\phi^{\prime}}(s^{\prime})+\epsilon,\ \epsilon\sim clip(\mathcal{N}(0,\tilde{\sigma}), -c,\ c)$ 
                    
                    \STATE Computes target: \\ $Q_{t} \leftarrow r+\gamma*\min_{i=1,2}Q_{\theta_i}(s', a')$
                    
                    
                    \STATE Update double critics with one step gradient descent:\\
                   $\nabla_{\theta_i} \frac{1}{N} \sum_i(Q_t - Q_{\theta_{i}(s_{t},a_{t})})^2$  \qquad for i=1,2
                    
                    \IF {t \% policy\_freq == 0}
                        \STATE Update policy with one step gradient descent:\\             
                        $\nabla_{\phi}\frac{1}{N} \sum_i[\nabla_{a_{t}}Q_{\theta_{1}}(s_{t},a_{t})\vert _{a_{t}=\pi(\phi)}\nabla_{\phi}\pi_{\phi}(s_{t})]$
                        
                        Soft update for the target networks: \\
                        \STATE $\phi^{\prime}\leftarrow\tau\phi+(1-\tau)\phi^{\prime}$
                        \STATE $\theta_{i}^{\prime}\leftarrow\tau\theta_{i}+(1-\tau)\theta_{i}^{\prime}$ \qquad for i=1,2

                    \ENDIF
                \ENDIF
            \ENDFOR
        \ENDFOR
  \end{algorithmic}
\end{algorithm}

We train for a number max steps ($max\_steps$) in a number episodes ($max\_steps$). Our approach starts by exploring random actions while a minimum number of steps ($start\_steps$) has not been achieved yet.

We use an LSTM as actor-network and denote them by $\phi$ and its copy $\phi^{\prime}$ as actor target. The double critic as well, by $\theta_{1}$, $\theta_{2}$ for parameterization of two value networks and its copies $\theta_{1}^{\prime}$, $\theta_{2}^{\prime}$ as critic targets. The learning of both networks happens simultaneously, addressing approximation error, reducing the bias, and finding the highest Q-value. The actor target chooses the action $a^{\prime}$ based on the state $s^{\prime}$, and we add Ornstein-Uhlenbeck noise to it. The double critic targets take the tuple ($s^{\prime}$, $a^{\prime}$) and return two Q-values as output. The minimum of the two target Q-values is considered as the approximated value return. The loss is calculated with the Mean Squared Error of the approximate value from the target networks and the value from the critic networks. We use Adaptive Moment Estimation (Adam) to minimize the loss.

We delay by updating the policy network less frequently than the value network. It is updated taking into account a $policy\_freq$ factor that increases over time by the following rule:

\begin{equation*} policy\_freq=(int)\frac{1}{0.5- \frac{t}{max\_steps \times 3}}. \end{equation*}

\subsection{Double Critic Deep Reinforcement Learning Stochastic - DoCRL-S}

A stochastic approach is also developed in this work. Our approach is based on the SAC algorithm~\cite{haarnoja2018soft}. This algorithm consists of a bias-stochastic actor-critic that combines off-policy updates with a stochastic actor-critic method to learn continuous action space policies. It uses neural networks as approximation functions to learn a policy and two Q-values functions similarly to TD3. However, SAC utilizes the current stochastic policy to act without noise, providing better stability and performance. It maximizes the reward and the policy's entropy, encouraging the agent to explore new states and improving training speed. We use the soft Bellman equation with neural networks as a function approximation to maximize the entropy. The pseudocode of DoCRL-S can be seen in Algorithm \ref{alg:docrl_s}.

\begin{algorithm}[!htb]
    \algsetup{linenosize=\tiny}
    \scriptsize
    \caption{Double Critic Deep Reinforcement Learning Stochastic - DoCRL-S}
    \label{alg:docrl_s}
    \begin{algorithmic}[1]
        \STATE Initialize critic networks $\theta_{1}$, $\theta_{2}$ , and actor network $\pi(\phi)$
        \STATE Initialize target networks $\phi^{\prime}\leftarrow\phi$, $\theta_{1}^{\prime}\leftarrow\theta_{1}$, $\theta_{2}^{\prime}\leftarrow\theta_{2}$
        \STATE Initialize replay buffer $\beta$
        \FOR{$ep = 1$ to $max\_eps$}
            \STATE reset environment state
            \FOR{$t = 0$ to $max\_steps$}
                \IF {$t < start\_steps$}
                    \STATE $a_{t}$ = env.action\_space.sample() 
                \ELSE
                    \STATE $a_t\leftarrow\pi_{\phi}(\cdot|s)$
                \ENDIF
                
                \STATE $s_{t+1}$, $r_{t}$, $d_{t}$, \_ env.step($a_{t}$)
                
                \STATE store the new transition $S$ $(s_{t}, a_{t}, r_{t}, s_{t+1}, d_{t})$ into $\beta$
                
                \IF{$t > start\_steps$}
                    \STATE Sample m-batch of $N$ transitions $(s_{t}, a_{t}, r_{t},s_{t+1}, d_{t})$ from $\beta$
                    
                    \STATE $double = ([min_{i=1,2}( Q_{\theta'_{i}}({s_{t}},{a_{t}}))-\alpha \log \pi ({a_{t}}|{s_{t}})])$
                    
                    \STATE $Q_t=r({s_{t}},{a_{t}})+\gamma(1-d_{t})*double$ 
                    
                    \STATE Update double critics with one step gradient descent:\\
                    $\nabla_{\theta_i} argmin_{\theta_{i}} \frac{1}{|N|} \sum(Q_t - Q_{\theta_{i}}(s_{t},a_{t}))^2$  \qquad for i=1,2
                    
                    \IF {t \% policy\_freq == 0}
                        
                        \STATE Update policy with one step gradient descent:\\
                        $ \nabla_{\phi} \frac{1}{|N|} \sum_{s_t \in \beta} ([min_{i=1,2}( Q_{\theta_{i}}({s_{t}},{a_{t_\phi}}))-\alpha \log \pi ({a_{t_\phi}}|{s_{t}})])$ \qquad for i=1,2

                        \STATE Soft update for the target networks:\\ $\theta_i^{\prime}\leftarrow\tau\theta_i+(1-\tau)\theta_i^{\prime}$ \qquad for i=1,2
                    \ENDIF
                \ENDIF
            \ENDFOR
        \ENDFOR
  \end{algorithmic}
\end{algorithm}

We train for a number max steps ($max\_steps$) in a number of episodes ($max\_steps$) as well. It explores random actions while a minimum number of steps ($start\_steps$) has not been achieved yet. A LSTM structure was used for the policy network $\phi$. After sampling a batch from the memory, we compute the targets for the Q-functions $Q_t({r_{t}},{s_{t+1}},{d_{t}})$, and update the Q-functions. We perform the delay by updating the policy less frequently than the value network, as performed with our DoCRL-D approach. It was updated taking into account the same $policy\_freq$ factor as the DoCRL-D. 

\subsection{Simulated Environments}

Our experiments are conducted with the vehicle and on the second and more complex environment provided by Grando et al. \cite{bedin2021deep}. Based on the Gazebo simulator together with ROS, it makes the use of the framework RotorS \cite{furrer2016rotors}, which provides the means to simulate aerial vehicles with different command levels, such as angular rates, attitude, location control and the simulation of wind with an Ornstein-Uhlenbeck noise method. The underwater simulation is enabled by the UUV simulator \cite{manhaes2016uuv}, which allows the simulation of hydrostatic and hydrodynamic effects, as well as thrusters, sensors, and external perturbations. With this framework, the vehicle's underwater model was defined with parameters such as the volume,  additional mass, center of buoyancy, etc., as well as the characteristics of the underwater environment itself.

The environment simulate a walled water tank, with dimensions of 10$\times$10$\times$6 meters and a one-meter water column. It has four cylindrical columns representing drilling risers.

\subsubsection{HUAUV Description} 


The vehicle used is based on the model presented by Drews-Jr \emph{et al.} \cite{drews2014hybrid}, Neto \emph{et al.} \cite{neto2015attitude} and \emph{et al.} \cite{horn2019study}. It was described using its actual mechanics settings, including the inertia, motor coefficients, mass, rotor velocity, and others. Its sensing was optimized for real-world LIDAR and Sonar. The described LIDAR is based on the UST 10LX model. It provides a 10 meters distance sensing with 270\degree range and 0.25\degree of resolution. It was simulated using the plugin ray of Gazebo. The simulated FLS sonar is based on the sonar simulation plugin developed by Cerqueira \emph{et al.}~\cite{cerqueira2017novel}. It has 20 meters of range, with a bin count of 1000 and a beam count of 256. The width and height angles of the beam were $90^0$ and $15^0$, respectively. 

\subsubsection{Network Structure and Rewarding System} \label{secapproach}
%




The structure of our approaches has a total of 26 dimensions for the state, 20 samples for the distance sensors, the three previous actions and three values related to the target goal, which are the vehicle's relative position to the target, and relative angles to the target in the x-y plane and the z-distance plan. When in the air, the 20 samples come from the LIDAR. We get these samples equally spaced by $13.5\degree$ in the $270\degree$ LIDAR. When underwater, the distance information comes from the Sonar. We also got 20 beams equally spaced among the total of 256, and we took for the highest bin in each beam. This conversion based on the range gives us the distance towards the obstacle or the tank's wall 
\cite{Santos18,Santos19}
.The actions are scaled between $0$ and $0.25$ $m/s$ for the linear velocity, from $-0.25$ $m/s$ to $0.25$ $m/s$ for the altitude velocity and from $-0.25$ to $0.25$ $rad$ for the $\Delta$ yaw.

\subsubsection{Reward Function} 

We proposed a binary rewarding function that takes into account a positive reward in case of success or a negative reward in case of failure or in case the episode ($ep$) ends at the 500 steps limit:

\vspace{-5mm}
\begin{equation}
r(s_t, a_t)= 
\begin{cases}
    r_{arrive}           & \text{if } d_t < c_d\\
    r_{collide}          & \text{if } min_x < c_o\ ||\ ep = 500\\
\end{cases}
\end{equation}

The reward $r_{arrive}$ is set to  100, while the negative reward $r_{collide}$ is set to -10. Both $c_d$ and $c_o$ distance were set to $0.5$ meters.
\section{Experimental Results}
\label{results} 

During the training phase, we created a randomly generated goal that the agent should navigate towards. The agents train for a maximum of 500 steps or until they collide with an obstacle or with the tank border. In case of reaching the goal before the limit of episodes, a new random goal was generated. In this case, the total amount of reward could exceed the maximum value of 100. It was used a learning rate of $10^{-3}$, a minibatch of 256 samples and the Adam optimizer for all approaches, including for the compared methods. We limited the number of episodes of 1500 episodes. These respective limits for the episode number ($max\_steps$) are used based on the stagnation of the maximum average reward received.

\subsection{Results} 

In this section, the results obtained during our evaluation are shown. An extensive amount of statistics are collected. The task addressed is goal-oriented navigation considering medium transition, where the robot must navigate from a starting point to an endpoint. This task was addressed in a two-fold way in our tests: starting in the air, performing the medium transition and navigating to a target underwater; and the other way around, starting underwater, performing the medium transition and navigating to a target in the air. We collected the statistics for each of our proposed models (DoCRL-D and DoCRL-S) and compared them with the performance of the state-of-the-art deterministic (Det.) and stochastic (Sto.) Deep-RL methods for HUAUVs 
, as well as a behavior-based algorithm \cite{marino2016minimalistic} 
. This two-fold task was performed for 100 trials and the total of successful trials are recorded. Also, the average time for both underwater ($t\_water$) and aerial ($t\_air$) along navigation with their standard deviations are recorded. 

\begin{table}[bp!]
\vspace{-5mm}
\centering
\setlength{\tabcolsep}{0.8pt}
\caption{Mean and standard deviation metrics over 100 navigation trials for all approaches.}
\label{table:mean_std}
\begin{tabular}{c c c c c} 
\toprule
Test & $t_{air}$ (s) & $t_{water}$ (s) & Success \\
\midrule
\textbf{A-W DoCRL-D} & $\textbf{14.55}$ $\pm$ $\textbf{0.87}$ & $11.19$ $\pm$ $2.86$ & $\textbf{100}$ \\
\textbf{A-W DoCRL-S} & $72.02$ $\pm$ $24.33$ & $\textbf{9.97}$ $\pm$ $\textbf{6.05}$ & $\textbf{70}$ \\
A-W Sto. Grando \emph{et al.} \cite{bedin2021deep} & $18.10$ $\pm$ $7.91$ & $1.08$ $\pm$ $2.38$ & $1$ \\
A-W Det. Grando \emph{et al.} \cite{bedin2021deep} & $34.09$ $\pm$ $23.85$ & $13.06$ $\pm$ $12.98$ & $14$ \\
A-W BBA & $34.70$ $\pm$ $30.22$ & $4.87$ $\pm$ $6.16$ & $11$ \\
\textbf{W-A DoCRL-D} & $\textbf{16.67}$ $\pm$ $\textbf{10.44}$ & $\textbf{4.65}$ $\pm$ $\textbf{0.69}$ & $\textbf{42}$ \\
\textbf{W-A DoCRL-S} & $\textbf{8.77}$ $\pm$ $\textbf{7.23}$ & $\textbf{5.98}$ $\pm$ $\textbf{1.90}$ & $\textbf{71}$ \\
W-A Sto. Grando \emph{et al.} \cite{bedin2021deep} & $15.33$ $\pm$ $23.75$ & $14.9$ $\pm$ $18.88$ & $12$ \\
W-A Det. Grando \emph{et al.} \cite{bedin2021deep} & $27.58$ $\pm$ $14.77$ & $12.65$ $\pm$ $7.89$ & $13$ \\
W-A BBA & $38.01$ $\pm$ $29.83$ & $3.79$ $\pm$ $0.27$ & $32$ \\

\bottomrule
\end{tabular}
\end{table}

We set the initial position for the Air-Water (A-W) trials to (0.0, 0.0, 2.5) in the Gazebo Cartesian coordinates for the three scenarios. The target position was set to (2.0, 3.0, -1.0) at the bottom of the water tank. For the Water-Air (W-A) evaluation, the coordinates were swapped. The same was done for the second and third scenarios, where the coordinates used were (0.0, 0.0, 2.5) and (3.6, -2.4, -1.0). The target was defined in a path with obstacles on the way.

\begin{figure*}[tp!]
\centering
 \subfloat[DoCRL-D, air-water. \label{fig:ddpg_2_air_to_water}]{
	\begin{minipage}[c][0.73\width]{0.241\textwidth}
	   \centering
	   \includegraphics[width=\textwidth]{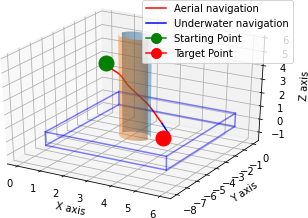}
	\end{minipage}}
 \hfill	
 \subfloat[DoCRL-D, air-water, x-y plane. \label{fig:ddpg_2_air_to_water_2d}]{
	\begin{minipage}[c][0.73\width]{0.241\textwidth}
	   \centering
	   \includegraphics[width=\textwidth]{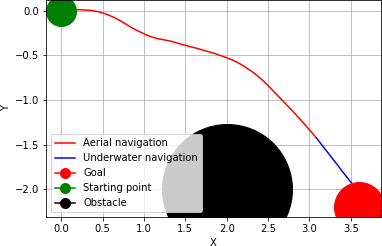}
	\end{minipage}}
 \hfill	
 \subfloat[DoCRL-D, water-air. \label{fig:ddpg_2_water_to_air}]{
	\begin{minipage}[c][0.73\width]{0.241\textwidth}
	   \centering
	   \includegraphics[width=\textwidth]{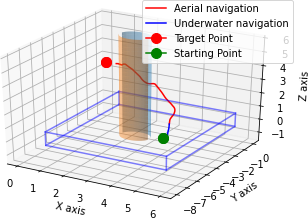}
	\end{minipage}}
 \hfill	
 \subfloat[DoCRL-D, water-air, x-y plane. \label{fig:ddpg_2_water_to_air_2d}]{
	\begin{minipage}[c][0.73\width]{0.241\textwidth}
	   \centering
	   \includegraphics[width=\textwidth]{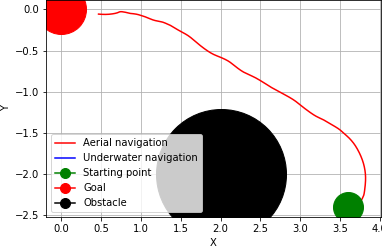}
	\end{minipage}}
 \hfill
 \subfloat[DoCRL-S, air-water. \label{fig:sac_2_air_to_water}]{
	\begin{minipage}[c][0.73\width]{0.241\textwidth}
	   \centering
	   \includegraphics[width=\textwidth]{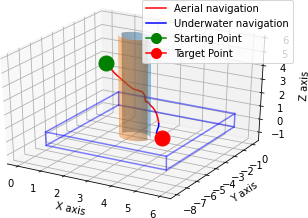}
	\end{minipage}}
 \hfill
 \subfloat[DoCRL-S, air-water, x-y plane. \label{fig:sac_2_air_to_water_2d}]{
	\begin{minipage}[c][0.73\width]{0.241\textwidth}
	   \centering
	   \includegraphics[width=\textwidth]{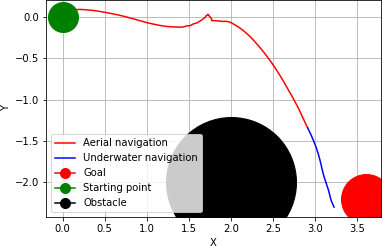}
	\end{minipage}}
 \hfill
 \subfloat[DoCRL-S, water-air. \label{fig:sac_2_water_to_air}]{
	\begin{minipage}[c][0.73\width]{0.241\textwidth}
	   \centering
	   \includegraphics[width=\textwidth]{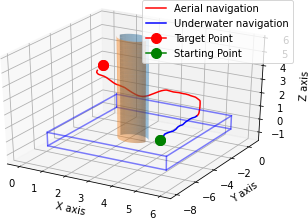}
	\end{minipage}}
 \hfill
 \subfloat[DoCRL-S, water-air, x-y plane. \label{fig:sac_2_water_to_air_2d}]{
	\begin{minipage}[c][0.73\width]{0.241\textwidth}
	   \centering
	   \includegraphics[width=\textwidth]{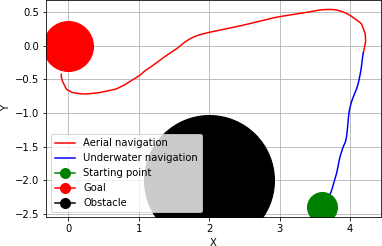}
	\end{minipage}}
 \hfill
 \subfloat[BBA, air-water. \label{fig:bug_2_air_to_water}]{
	\begin{minipage}[c][0.73\width]{0.241\textwidth}
	   \centering
	   \includegraphics[width=\textwidth]{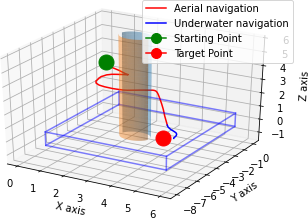}
	\end{minipage}}
 \hfill
 \subfloat[BBA, air-water, x-y plane. \label{fig:bug_2_air_to_water_2d}]{
	\begin{minipage}[c][0.73\width]{0.241\textwidth}
	   \centering
	   \includegraphics[width=\textwidth]{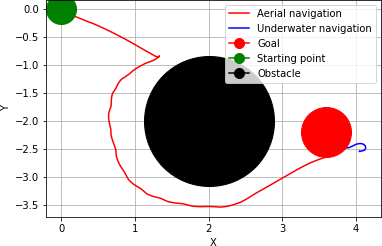}
	\end{minipage}}
 \hfill
 \subfloat[BBA, water-air. \label{fig:bug_2_water_to_air}]{
	\begin{minipage}[c][0.73\width]{0.241\textwidth}
	   \centering
	   \includegraphics[width=\textwidth]{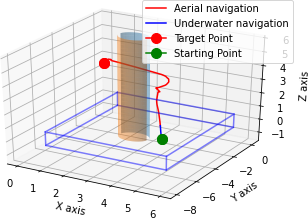}
	\end{minipage}}
 \hfill
 \subfloat[BBA, water-air, x-y plane. \label{fig:bug_2_water_to_air_2d}]{
	\begin{minipage}[c][0.73\width]{0.241\textwidth}
	   \centering
	   \includegraphics[width=\textwidth]{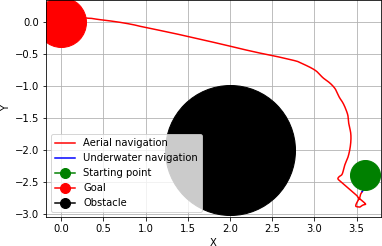}
	\end{minipage}}
 \hfill
\caption{Behavior sample of each approach tested in the environment trained. (\cref{fig:ddpg_2_air_to_water,fig:ddpg_2_air_to_water_2d,fig:ddpg_2_water_to_air,fig:ddpg_2_water_to_air_2d,fig:sac_2_air_to_water,fig:sac_2_air_to_water_2d,fig:sac_2_water_to_air,fig:sac_2_water_to_air_2d,fig:bug_2_air_to_water,fig:bug_2_air_to_water_2d,fig:bug_2_water_to_air,fig:bug_2_water_to_air_2d}).}
\label{fig:traj_nav}
\vspace{-6mm}
\end{figure*}

\section{Discussion}
\label{discussion}


The evaluation shows an overall increase in performance in navigation through the the scenario.  We can see that the DoCRL-D approach achieves a consistent performance of 100 successfully air-to-water navigation trials with also a consistent navigation time ($14.55 \pm 0.87$ and $11.19 \pm 2.86$). In this same scenario, the DoCRL-S approach performed a little worse in air-to-water navigation but outperformed the deterministic approach in water-to-air navigation. 


It is important to mention that these approaches are extensively evaluated in a realistic simulation, including control issues and disturbances such as wind. Thus, the results indicate that our approach may achieve real-world application if the correct data from the sensing and the relative localization are correctly ensured. 

\section{Conclusions}
\label{conclusion}

In this paper, we presented two novels Deep-RL approaches based on RNNs and double critic structures to perform the navigation of a HUAUV. By using physically realistic simulation in several water-tank-based scenarios, we showed that our approaches achieved an overall better capability to perform autonomous navigation, obstacle avoidance and medium transition than other approaches. Disturbances such as wind were successfully assimilated and good generalization through different scenarios was achieved. With our simple and realistic sensing approach that took into account only the range information, we presented overall better performance than the state-of-the-art and a classical behavior-like algorithm. Our work aims to contribute to future studies focusing on rescue missions in robotics, particularly with vehicles like the HUAUV, and extending to environmental rescue tasks in both air and water domains. Furthermore, we are actively conducting research to use our HUAUV model to explore its potential in practical real world applications.

\section*{Acknowledgment}


The authors would like to thank the VersusAI team. This work was partly supported by the CAPES, CNPq and PRH-ANP.

\vspace{-2mm}
\bibliographystyle{./bibliography/IEEEtran}
\bibliography{./bibliography/IEEEabrv,./bibliography/IEEEexample}

\begin{thebibliography}{10}
\providecommand{\url}[1]{#1}
\csname url@samestyle\endcsname
\providecommand{\newblock}{\relax}
\providecommand{\bibinfo}[2]{#2}
\providecommand{\BIBentrySTDinterwordspacing}{\spaceskip=0pt\relax}
\providecommand{\BIBentryALTinterwordstretchfactor}{4}
\providecommand{\BIBentryALTinterwordspacing}{\spaceskip=\fontdimen2\font plus
\BIBentryALTinterwordstretchfactor\fontdimen3\font minus
  \fontdimen4\font\relax}
\providecommand{\BIBforeignlanguage}[2]{{%
\expandafter\ifx\csname l@#1\endcsname\relax
\typeout{** WARNING: IEEEtran.bst: No hyphenation pattern has been}%
\typeout{** loaded for the language `#1'. Using the pattern for}%
\typeout{** the default language instead.}%
\else
\language=\csname l@#1\endcsname
\fi
#2}}
\providecommand{\BIBdecl}{\relax}
\BIBdecl

\bibitem{drews2014hybrid}
P.~L. Drews-Jr, A.~A. Neto, and M.~F. Campos, ``Hybrid unmanned aerial
  underwater vehicle: Modeling and simulation,'' in \emph{IEEE/RSJ IROS}, 2014,
  pp. 4637--4642.

\bibitem{neto2015attitude}
A.~A. Neto, L.~A. Mozelli, P.~L. Drews-Jr, and M.~F. Campos, ``Attitude control
  for an hybrid unmanned aerial underwater vehicle: A robust switched strategy
  with global stability,'' in \emph{IEEE ICRA}, 2015, pp. 395--400.

\bibitem{da2018comparative}
R.~T. da~Rosa, P.~J. Evald, P.~L. Drews-Jr, A.~A. Neto, A.~C. Horn, R.~Z.
  Azzolin, and S.~S. Botelho, ``A comparative study on sigma-point kalman
  filters for trajectory estimation of hybrid aerial-aquatic vehicles,'' in
  \emph{IEEE/RSJ IROS}, 2018, pp. 7460--7465.

\bibitem{maia2017design}
M.~M. Maia, D.~A. Mercado, and F.~J. Diez, ``Design and implementation of
  multirotor aerial-underwater vehicles with experimental results,'' in
  \emph{IEEE/RSJ IROS}, 2017, pp. 961--966.

\bibitem{lu2019multimodal}
D.~Lu, C.~Xiong, Z.~Zeng, and L.~Lian, ``A multimodal aerial underwater vehicle
  with extended endurance and capabilities,'' in \emph{IEEE ICRA}, 2019, pp.
  4674--4680.

\bibitem{mercado2019aerial}
D.~Mercado, M.~Maia, and F.~J. Diez, ``Aerial-underwater systems, a new
  paradigm in unmanned vehicles,'' \emph{Journal of Intelligent \& Robotic
  Systems}, vol.~95, no.~1, pp. 229--238, 2019.

\bibitem{horn20}
A.~C. Horn, P.~M. Pinheiro, R.~B. Grando, C.~B. da~Silva, A.~A. Neto, and P.~L.
  Drews-Jr, ``A novel concept for hybrid unmanned aerial underwater vehicles
  focused on aquatic performance,'' in \emph{IEEE LARS/SBR}, 2020, pp. 1--6.

\bibitem{aoki2021}
V.~M. Aoki, P.~M. Pinheiro, P.~L.~J. Drews-Jr, M.~A.~B. Cunha, and L.~G.
  Tuchtenhagen, ``Analysis of a hybrid unmanned aerial underwater vehicle
  considering the environment transition,'' in \emph{IEEE LARS/SBR}, 2021, pp.
  90--95.

\bibitem{bedin2021deep}
R.~B. Grando, J.~C. de~Jesus, V.~A. Kich, A.~H. Kolling, N.~P. Bortoluzzi,
  P.~M. Pinheiro, A.~Alves~Neto, and P.~L.~J. Drews-Jr, ``Deep reinforcement
  learning for mapless navigation of a hybrid aerial underwater vehicle with
  medium transition,'' in \emph{IEEE ICRA}, 2021, pp. 1088--1094.

\bibitem{ota2020efficient}
K.~Ota, Y.~Sasaki, D.~K. Jha, Y.~Yoshiyasu, and A.~Kanezaki, ``Efficient
  exploration in constrained environments with goal-oriented reference path,''
  in \emph{IEEE/RSJ IROS}, 2020, pp. 6061--6068.

\bibitem{tong2021uav}
G.~Tong, N.~Jiang, L.~Biyue, Z.~Xi, W.~Ya, and D.~Wenbo, ``{{UAV}} navigation
  in high dynamic environments: A deep reinforcement learning approach,''
  \emph{Chinese Journal of Aeronautics}, vol.~34, no.~2, pp. 479--489, 2021.

\bibitem{grando2022double}
R.~B. Grando, J.~C. de~Jesus, V.~A. Kich, A.~H. Kolling, and P.~L.~J. Drews-Jr,
  ``Double critic deep reinforcement learning for mapless 3d navigation of
  unmanned aerial vehicles,'' \emph{Journal of Intelligent \& Robotic Systems},
  vol. 104, no.~2, pp. 1--14, 2022.

\bibitem{carlucho2018}
I.~{Carlucho}, M.~{De Paula}, S.~{Wang}, B.~V. {Menna}, Y.~R. {Petillot}, and
  G.~G. {Acosta}, ``Auv position tracking control using end-to-end deep
  reinforcement learning,'' in \emph{MTS/IEEE OCEANS}, 2018.

\bibitem{fujimoto2018addressing}
S.~Fujimoto, H.~Hoof, and D.~Meger, ``Addressing function approximation error
  in actor-critic methods,'' in \emph{ICML}, 2018, pp. 1587--1596.

\bibitem{haarnoja2018soft}
T.~Haarnoja, A.~Zhou, P.~Abbeel, and S.~Levine, ``Soft actor-critic: Off-policy
  maximum entropy deep reinforcement learning with a stochastic actor,'' in
  \emph{ICML}, vol.~80, 2018, pp. 1861--1870.

\bibitem{marino2016minimalistic}
R.~Marino, F.~Mastrogiovanni, A.~Sgorbissa, and R.~Zaccaria, ``A minimalistic
  quadrotor navigation strategy for indoor multi-floor scenarios,'' in
  \emph{Intelligent Autonomous Systems 13}.\hskip 1em plus 0.5em minus
  0.4em\relax Springer, 2016, pp. 1561--1570.

\bibitem{tobin2017domain}
J.~Tobin, R.~Fong, A.~Ray, J.~Schneider, W.~Zaremba, and P.~Abbeel, ``Domain
  randomization for transferring deep neural networks from simulation to the
  real world,'' in \emph{IEEE/RSJ IROS}, 2017.

\bibitem{tai2017virtual}
L.~Tai, G.~Paolo, and M.~Liu, ``Virtual-to-real deep reinforcement learning:
  Continuous control of mobile robots for mapless navigation,'' in
  \emph{IEEE/RSJ IROS}, 2017, pp. 31--36.

\bibitem{jesus2019deep}
J.~C. de~Jesus, J.~A. Bottega, M.~A. Cuadros, and D.~F. Gamarra, ``Deep
  deterministic policy gradient for navigation of mobile robots in simulated
  environments,'' in \emph{19th ICAR}, 2019, pp. 362--367.

\bibitem{jesus2021soft}
J.~C. de~Jesus, V.~A. Kich, A.~H. Kolling, R.~B. Grando, M.~A. d. S.~L.
  Cuadros, and D.~F.~T. Gamarra, ``Soft actor-critic for navigation of mobile
  robots,'' \emph{Journal of Intelligent \& Robotic Systems}, vol. 102, no.~2,
  pp. 1--11, 2021.

\bibitem{kelchtermans2017hard}
K.~Kelchtermans and T.~Tuytelaars, ``How hard is it to cross the
  room?--training (recurrent) neural networks to steer a {UAV},'' \emph{arXiv
  preprint arXiv:1702.07600}, 2017.

\bibitem{singh2018mobile}
N.~H. Singh and K.~Thongam, ``Mobile robot navigation using mlp-bp approaches
  in dynamic environments.'' \emph{Arabian Journal for Science \& Engineering},
  vol.~43, p. 8013–8028, 2018.

\bibitem{grando2020visual}
R.~B. Grando, P.~M. Pinheiro, N.~P. Bortoluzzi, C.~B. da~Silva, O.~F. Zauk,
  M.~O. Piñeiro, V.~M. Aoki, A.~L. Kelbouscas, Y.~B. Lima, P.~L. Drews-Jr, and
  A.~A. Neto, ``Visual-based autonomous unmanned aerial vehicle for inspection
  in indoor environments,'' in \emph{IEEE LARS/SBR}, 2020, pp. 1--6.

\bibitem{rodriguez2018deep}
A.~Rodriguez-Ramos, C.~Sampedro, H.~Bavle, I.~G. Moreno, and P.~Campoy, ``A
  deep reinforcement learning technique for vision-based autonomous multirotor
  landing on a moving platform,'' in \emph{IEEE/RSJ IROS}, 2018, pp.
  1010--1017.

\bibitem{furrer2016rotors}
F.~Furrer, M.~Burri, M.~Achtelik, and R.~Siegwart, ``Rotors—a modular gazebo
  mav simulator framework,'' in \emph{Robot Operating System (ROS)}, 2016, pp.
  595--625.

\bibitem{sampedro2019fully}
C.~Sampedro, A.~Rodriguez-Ramos, H.~Bavle, A.~Carrio, P.~de~la Puente, and
  P.~Campoy, ``A fully-autonomous aerial robot for search and rescue
  applications in indoor environments using learning-based techniques,''
  \emph{Journal of Intelligent \& Robotic Systems}, pp. 601--627, 2019.

\bibitem{kang2019generalization}
K.~Kang, S.~Belkhale, G.~Kahn, P.~Abbeel, and S.~Levine, ``Generalization
  through simulation: Integrating simulated and real data into deep
  reinforcement learning for vision-based autonomous flight,'' in \emph{IEEE
  ICRA}, 2019, pp. 6008--6014.

\bibitem{2020arXiv201002293M}
G.~M. Barros and E.~L. Colombini, ``{Using Soft Actor-Critic for Low-Level
  {UAV} Control},'' \emph{arXiv e-prints}, vol. abs/2010.02293, 2020.

\bibitem{grando2020deep}
R.~B. Grando, J.~C. de~Jesus, and P.~L. Drews-Jr, ``Deep reinforcement learning
  for mapless navigation of unmanned aerial vehicles,'' in \emph{IEEE
  LARS/SBR}, 2020, pp. 1--6.

\bibitem{pinheiro2021trajectory}
P.~M. Pinheiro, A.~A. Neto, R.~B. Grando, C.~B. da~Silva, V.~M. Aoki,
  D.~Cardoso, A.~C. Horn, and P.~L.~J. Drews-Jr, ``Trajectory planning for
  hybrid unmanned aerial underwater vehicles with smooth media transition,''
  \emph{arXiv preprint arXiv:2112.13819}, 2021.

\bibitem{mnih2013playing}
V.~Mnih, K.~Kavukcuoglu, D.~Silver, A.~Graves, I.~Antonoglou, D.~Wierstra, and
  M.~A. Riedmiller, ``Playing atari with deep reinforcement learning,''
  \emph{NIPS Deep Learning Workshop}, 2013.

\bibitem{lillicrap2015continuous}
T.~P. Lillicrap, J.~J. Hunt, A.~Pritzel, N.~Heess, T.~Erez, Y.~Tassa,
  D.~Silver, and D.~Wierstra, ``Continuous control with deep reinforcement
  learning,'' in \emph{ICLR}, 2016.

\bibitem{manhaes2016uuv}
M.~M.~M. Manh{\~a}es, S.~A. Scherer, M.~Voss, L.~R. Douat, and T.~Rauschenbach,
  ``{UUV} simulator: A gazebo-based package for underwater intervention and
  multi-robot simulation,'' in \emph{MTS/IEEE OCEANS}, 2016, pp. 1--8.

\bibitem{horn2019study}
A.~C. Horn, P.~M. Pinheiro, C.~B. Silva, A.~A. Neto, and P.~L. Drews-Jr, ``A
  study on configuration of propellers for multirotor-like hybrid
  aerial-aquatic vehicles,'' in \emph{19th ICAR}, 2019, pp. 173--178.

\bibitem{cerqueira2017novel}
R.~Cerqueira, T.~Trocoli, G.~Neves, S.~Joyeux, J.~Albiez, and L.~Oliveira, ``A
  novel gpu-based sonar simulator for real-time applications,'' \emph{Computers
  \& Graphics}, vol.~68, pp. 66--76, 2017.

\bibitem{Santos18}
M.~M. Santos, P.~Drews-Jr, P.~N{\'u}{\~n}ez, and S.~Botelho, ``Object
  recognition and semantic mapping for underwater vehicles using sonar data,''
  \emph{Journal of Intelligent \& Robotic Systems}, vol.~91, no.~2, pp.
  279--289, 2018.

\bibitem{Santos19}
M.~M. Santos, G.~B. Zaffari, P.~O. C.~S. Ribeiro, P.~L.~J. Drews-Jr, and
  S.~S.~C. Botelho, ``Underwater place recognition using forward-looking sonar
  images: A topological approach,'' \emph{Journal of Field Robotics}, vol.~36,
  no.~2, pp. 355--369, 2019.

\end{thebibliography}

\end{document}